\documentclass[runningheads]{llncs}

\usepackage[mobile]{eccv}

\usepackage{eccvabbrv}
\usepackage{wrapfig}
\usepackage{graphicx}
\usepackage{booktabs}
\usepackage{subcaption}
\usepackage{epsfig}
\usepackage{amsmath}
\usepackage{amssymb}
\usepackage{epigraph}
\usepackage{bm}
\usepackage{multirow}
\usepackage[misc]{ifsym}
\usepackage{epsfig}
\usepackage{epigraph}

\makeatletter
\newcommand{\rmnum}[1]{\romannumeral #1}
\newcommand{\Rmnum}[1]{\expandafter\@slowromancap\romannumeral #1@}
\makeatother
\usepackage{colortbl}
\usepackage[accsupp]{axessibility}

\usepackage[pagebackref,breaklinks,colorlinks,citecolor=eccvblue]{hyperref}
\usepackage{orcidlink}

\begin{document}

% ---------------------------------------------------------------
% TODO REVIEW: Replace with your title
\title{Towards Multimodal Sentiment Analysis Debiasing via Bias Purification} 

% TODO REVIEW: If the paper title is too long for the running head, you can set
% an abbreviated paper title here. If not, comment out.
\titlerunning{Multimodal Counterfactual Inference Sentiment Analysis}

\author{
Dingkang Yang\inst{1,3}$^{\ast \dagger}$ \and
Mingcheng Li\inst{1,3}$^{\ast}$\and
Dongling Xiao\inst{2}  \and
Yang Liu\inst{1}  \and
Kun Yang\inst{1}  \and
Zhaoyu Chen\inst{1}  \and
Yuzheng Wang\inst{1}  \and
Peng Zhai\inst{1}  \and
Ke Li\inst{2} \and
Lihua Zhang\inst{1,3,4}$^{\textrm{\Letter}}$
}

\authorrunning{D. Yang et al.}

\institute{Academy for Engineering \& Technology, Fudan University \and
Tencent Youtu Lab  \and
Cognition and Intelligent Technology Laboratory (CIT Lab)  \and
Engineering Research Center of AI and Robotics, Ministry of Education, China \\
\email{dkyang20@fudan.edu.cn, mingchengli21@m.fudan.edu.cn} 
}

\renewcommand{\thefootnote}{}
\footnotetext{$^{\ast}$Equal contribution. $^{\textrm{\Letter}}$Corresponding author.}
\footnotetext{$^{\dagger}$Work done during the internship at Tencent Youtu Lab.}

\maketitle

\begin{abstract}
Multimodal Sentiment Analysis (MSA) aims to understand human intentions by integrating emotion-related clues from diverse modalities, such as visual, language, and audio.
Unfortunately, the current MSA task invariably suffers from unplanned dataset biases, particularly multimodal utterance-level label bias and word-level context bias. These harmful biases potentially mislead models to focus on statistical shortcuts and spurious correlations, causing severe performance bottlenecks. To alleviate these issues, we present a Multimodal Counterfactual Inference Sentiment (MCIS) analysis framework based on causality rather than conventional likelihood. 
Concretely, we first formulate a causal graph to discover harmful biases from already-trained vanilla models.
In the inference phase, given a factual multimodal input, MCIS imagines two counterfactual scenarios to purify and mitigate 
these biases. Then, MCIS can make unbiased decisions from biased observations by comparing factual and counterfactual outcomes. We conduct extensive experiments on several standard MSA benchmarks. Qualitative and quantitative results show the effectiveness of the proposed framework.
\end{abstract}

\section{Introduction}
\label{sec:intro}
\epigraph{\emph{``Believe nothing you hear, and only one half that you see.''}}{\scriptsize-\emph{Edgar Allan Poe, The System of Doctor Tarr and Professor Fether}}

As an essential task in human intention understanding, Multimodal Sentiment Analysis (MSA) \cite{zadeh2016multimodal, zadeh2018multimodal,li2024correlation,li2024unified} attempts to empower machines with the senses of ``hearing'' \cite{kshirsagar2022quality} and ``seeing'' \cite{lu2022domain} to mimic human perception of emotions from diverse modalities.
Following the traditional likelihood rule, most existing studies focus on improving MSA performance by exploiting various strategies, including disentangled representation learning \cite{tsai2018learning,Hazarika2020misa, yang2022disentangled,li2023decoupled}, attention-based cross-modal interactions \cite{tsai2019multimodal,liang2021attention,lv2021progressive,yang2022contextual,yang2022learning,lei2023text,yang2024towards}, fusion mechanisms \cite{pham2019found,zadeh2017tensor,liu2018efficient,rahman2020integrating,sun2020learning,yang2022emotion,yang2023target}, and well-designed auxiliary tasks \cite{yu2021learning,wu2021text,li2023towards}. Despite the impressive improvements achieved by numerous works, they all invariably captured harmful dataset biases \cite{sun2022counterfactual, niu2021counterfactual, feder2021causalm} and suffered from unintended confounders \cite{tian2022debiasing,qian2021counterfactual,yang2023context}, which are multimodal utterance-level  \textbf{label bias} and word-level \textbf{context bias}.
\begin{figure}[t]
  \centering
  \includegraphics[width=0.65\linewidth]{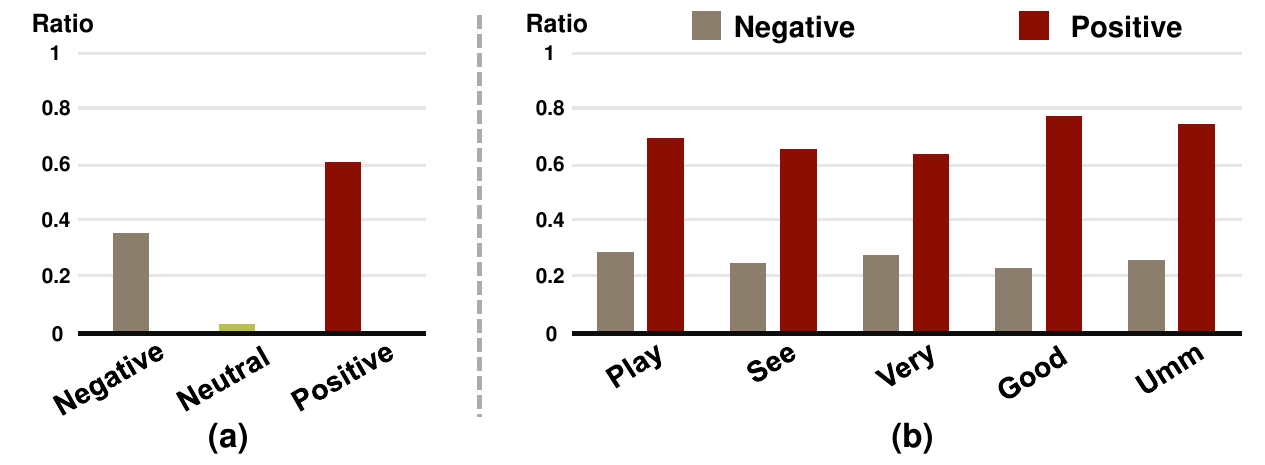}
  \caption{The distribution of (a) sentiment labels and (b) several context words from the training set on the MOSI dataset \cite{zadeh2016multimodal}.
  }
  \label{intro1}
\end{figure}

The harmful label bias usually occurs when the number of training samples for a specific category is more significant than for other categories. For instance, \cref{intro1}(a) illustrates that the positive samples dominate MOSI dataset \cite{zadeh2016multimodal} compared to the other samples. Worse still, a binary sentiment analysis dataset could have a label distribution of 95\% : 5\% \cite{dixon2018measuring}. In this case, many previous studies \cite{zhang2014explicit, qian2021counterfactual,dixon2018measuring} have indicated that such unbalanced data distribution would lead to trained models relying heavily on label bias as \emph{statistical shortcuts} to make inaccurate predictions. Different from unimodal tasks that potentially convey the adverse effects via specific modalities~\cite{wang2020visual,qian2021counterfactual}, most MSA models are poisoned with side effects captured by multimodal representations due to multiple modalities in each sample sharing the same sentiment label \cite{busso2008iemocap, zadeh2016multimodal, zadeh2018multimodal}.

\begin{figure}[t]
  \centering
  \includegraphics[width=\linewidth]{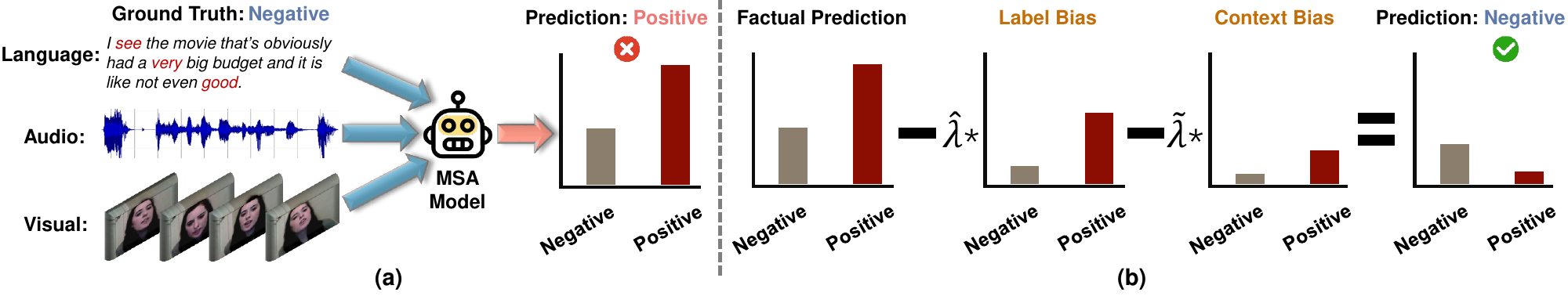}
  \caption{An example of multimodal sentiment analysis. (a) Likelihood-based biased prediction from re-implemented model DMD \cite{li2023decoupled}. (b) Unbiased prediction from the same model in the proposed framework. Binary classification results for illustration.
  }
  \label{intro2}
\end{figure}

Moreover, previous studies \cite{tsai2019multimodal, wu2021text, Hazarika2020misa} have demonstrated that language modality plays an important role in MSA compared to non-linguistic modalities, \ie, a suitable language model could achieve considerable performance \cite{pham2019found}. Nevertheless, linguistic information is not always beneficial due to the inherent context bias \cite{lin2015don,qian2021counterfactual}. The fatal context bias generally emerges when trained models exhibit strong \textit{spurious correlations} between specific categories and context words in language modality. In \cref{intro1}(b), some emotionally ambiguous words appear with imbalanced frequency in negative and positive samples. Consequently, MSA models tend to predict samples containing those words to an incorrect category based on biased statistical information rather than intrinsic textual semantics \cite{waseem2016hateful, qian2021counterfactual}. For example, \cref{intro2}(a) shows the predicted binary classification result from a state-of-the-art (SOTA) model \cite{li2023decoupled} on the MOSI. As the context words ``good'' and ``very'' appear more frequently in the positive than in the negative samples in the training set, the model predicts the testing sample as ``positive'' via an unreliable association. Therefore, to perform more reasonable sentiment inference, we need to suitably purify and eliminate the prejudicial effects caused by these biases in prior observations, as shown in \cref{intro2}(b).

Unlike machines that make biased predictions directly from an inference process by considering prior observations, humans have a natural counterfactual intuition~\cite{niu2021counterfactual}. Specifically, even though we are born and learn in a biased world, the counterfactual ability \cite{tang2020unbiased} enables us to make unbiased decisions by removing exogenous interference (\eg, label bias under limited observations) and endogenous reason (\eg, language context bias). The underlying mechanism is causality-based: decisions are made by counterfactual inference to pursue a true causal effect rather than a statistical shortcut or spurious correlation. To this end, we depict the counterfactual scenario as follows:

\noindent \textbf{Counterfactual MSA:} \textit{What will the prediction be, if the model does not see the multimodal input or only sees context words in the language modality?}

Intuitively, the counterfactual MSA have two outcomes: (1) the trained model relies purely on the statistical shortcut for prediction under the no-treatment condition of the multimodal input. In this case, \cref{intro2}(b) shows that the purified label bias results in a higher probability of ``positive'' than ``negative''. 
(2) The trained model relies only on the spurious correlation for prediction under the intervention of preserving context words solely. The result contains the pure side effect obtained by distilling the context bias. Motivated by the above observations, we propose a Multimodal Counterfactual Inference Sentiment (MCIS) analysis framework to mitigate the deleterious impact of two types of dataset biases. Concretely, we first design a tailored causal graph for MSA to diagnose causalities among variables and identify the dataset biases as unintended confounders.
The proposed framework is \textit{parameter-free} and \textit{training-free}, meaning that MCIS accommodates already-trained models following biased vanilla training via our generalized causal graph.
During the inference phase, MCIS intervenes with confounding multimodal inputs via backdoor adjustment theory \cite{pearl2009causal,qian2021counterfactual} to mimic the two counterfactual outcomes described above. By subtracting the counterfactual outcomes of the pure dataset biases, MCIS consistently improves the performance of SOTA models with unbiased predictions.

The main contributions are summarized as follows:
\begin{itemize}
\item We are the first to identify and disentangle the label and context biases in the MSA task from a novel causal inference perspective.
Based on innate human counterfactual intuition, we empower models to achieve unbiased predictions in biased observations.
\item Our causality-based MCIS is general and suitable for different MSA architectures and fusion mechanisms.
\item  Comprehensive experiments on several MSA benchmarks demonstrate the effectiveness of our framework.
\end{itemize}
\section{Related Work}
\label{sec:related}

\noindent \textbf{Multimodal Sentiment Analysis.}
Instead of modeling linguistic information alone \cite{ poria2016sentic}, MSA aims to integrate additional non-linguistic modalities to learn sentiment-related representations, such as visual \cite{zadeh2018multimodal} and acoustic signals \cite{yang2022contextual}. Driven by learning-based techniques~\cite{yang2023how2comm,yang2023aide,yang2023what2comm,chen2024miss,kuang2023towards,liu2023generalized}, mainstream MSA studies follow two aspects: representation learning and multimodal fusion. Multimodal representation learning \cite{Hazarika2020misa,liang2021attention,yu2020ch,tsai2018learning,wu2021text,li2023decoupled} attends to mitigating modality gap
or information redundancy to obtain refined modality semantics. For instance,  Hazarika \etal \cite{Hazarika2020misa} advocated projecting each modality into modality-invariant and -specific spaces to learn complementary information.  
For multimodal fusion, previous works \cite{lv2021progressive, tsai2019multimodal, zadeh2017tensor, rahman2020integrating, liu2018efficient} have explored sophisticated fusion strategies and mechanisms to obtain effective representations.
As a typical example, Tsai \etal \cite{tsai2019multimodal} achieved potential adaption fusion from one modality to another based on multimodal transformers. 
Despite the impressive improvements achieved by previous studies following traditional likelihood estimation, they invariably ignored the adverse effects of the dataset biases, resulting in biased predictions. 
In comparison, we achieve unbiased decisions by exploiting causality-based counterfactual thinking. 
The proposed framework significantly improves the performance of existing models without \textit{any complex network designs and parameters}.

\noindent \textbf{Causal Inference.}
Causal inference is a tool that seeks actual effects in a specific phenomenon \cite{pearl2009causal}. Currently, the mainstream causal inference studies applied to deep learning consist of two aspects: intervention \cite{wang2020visual,chen2022causal,tian2022debiasing,yang2024suci} and counterfactuals \cite{niu2021counterfactual,tang2020unbiased,sun2022counterfactual,qian2021counterfactual,yang2024robust}. \emph{Intervention} is an operation that alters original data distribution to discover causal effects \cite{glymour2016causal}. \emph{Counterfactuals} depict imagined outcomes produced by factual variables under different treatments \cite{pearl2009causality}. Our study focuses on obtaining counterfactual outcomes via intervention. Causal inference can remove confounders in data and learn actual causal effects instead of spurious associations, so it has been widely used in many downstream tasks to improve the models'  performance, including visual question answer \cite{niu2021counterfactual}, natural language understanding \cite{tian2022debiasing}, and scene graph generation \cite{tang2020unbiased}.
A recent study~\cite{sun2022counterfactual} focused on designing an additional model to capture the harmful effect of textual semantics.
However, they ignored the label bias and failed to disentangle the main content and context at the word level, thus incapable of language bias ascription.
Different from previous efforts~\cite{qian2021counterfactual,sun2022counterfactual}, this is the first work to identify both label bias and context bias in the MSA task from a causal perspective. 
Our framework effectively eliminates the side effects of dataset biases from multimodal inputs, which makes a step towards unbiased prediction in this field.

\section{Methodology}
\subsection{Framework Overview}
\begin{figure*}[t]
  \centering
  \includegraphics[width=\textwidth]{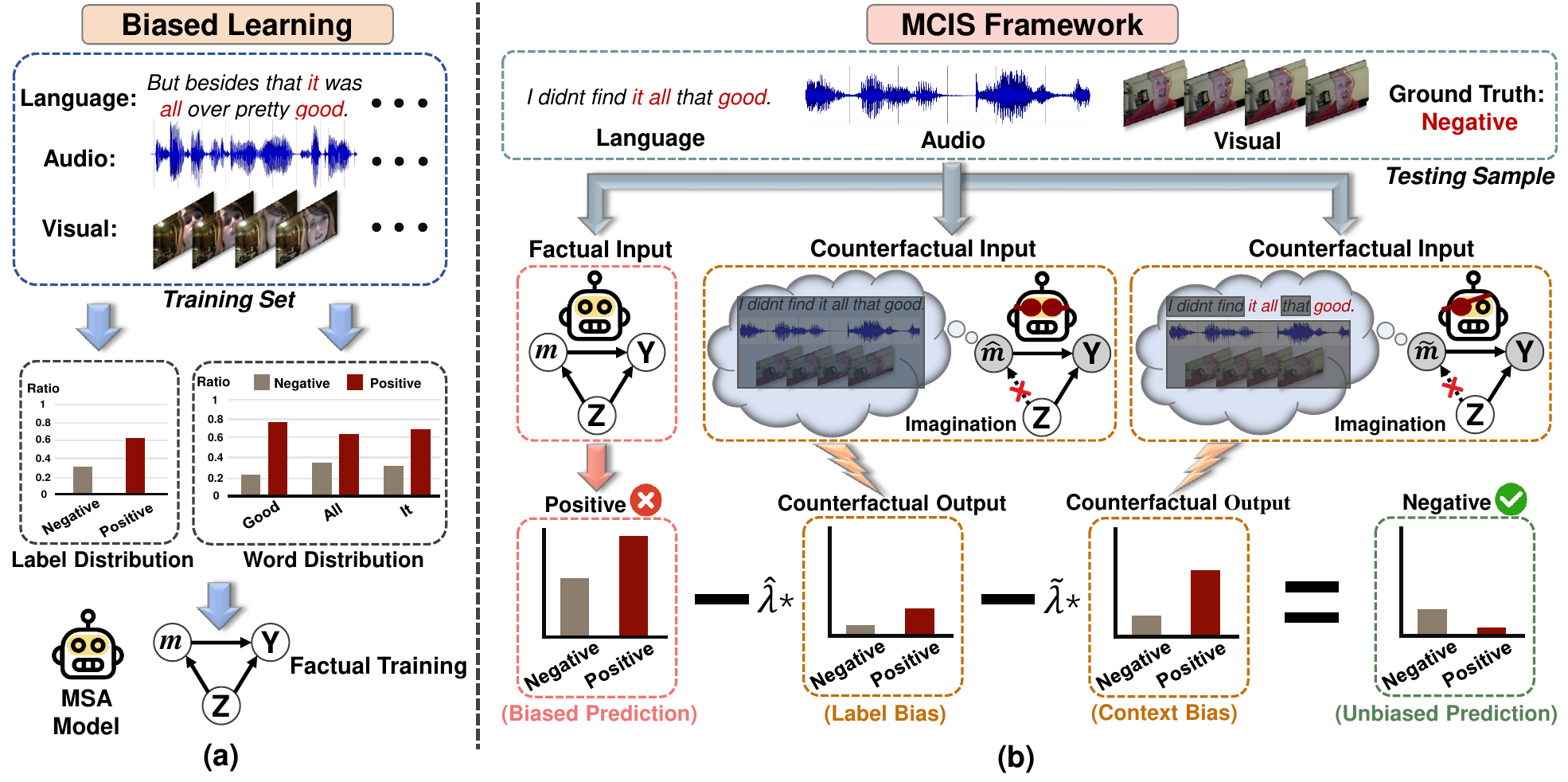}
  \caption{ (a) The biased learning of MSA models follows the factual training. (b) The architecture of our MCIS framework. MCIS compares factual and counterfactual outcomes for different multimodal input treatments. By subtracting the label and context biases, MCIS can achieve unbiased predictions from biased observations.
 }
  \label{model}
\end{figure*}

The proposed MCIS framework is illustrated in \cref{model}(b).
Concretely, 
MCIS allows already-trained models to preserve harmful dataset biases via biased conventional learning.
Given a factual multimodal input in the inference phase, MCIS imagines two types of multimodal counterfactual inputs to obtain two counterfactual outputs: purified label bias and context bias. Eventually, MCIS performs a bias elimination strategy in adaptive proportions to obtain unbiased counterfactual predictions by comparing factual and counterfactual outcomes.
 \begin{figure}[t]
  \centering
  \includegraphics[width=\linewidth]{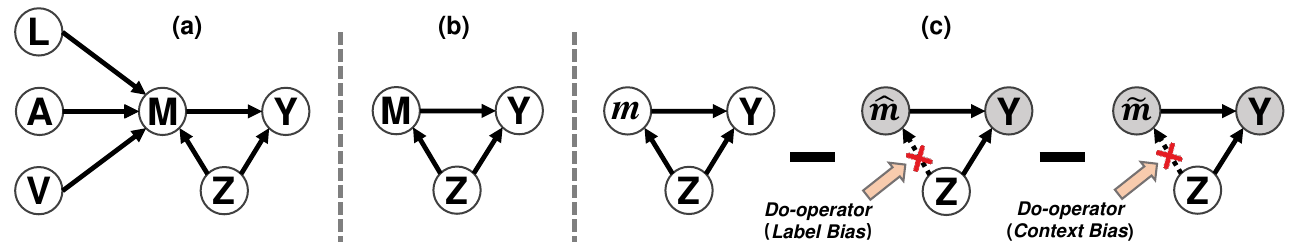}
  \caption{(a) The tailored causal graph for MSA. (b) The simplified causal graph for MSA. (c) Comparison between factual MSA and counterfactual MSA. White nodes are at the value $M = m$ while gray nodes are at the value $M = \hat{m} $ or $M = \tilde{m} $.
  }
  \label{graph}
\end{figure}

\subsection{Structural Causal Graph in MSA}

\noindent \textbf{Problem Formalization.}
Given multimodal utterance inputs from video segments, MSA aims to predict sentiment scores by learning multimodal models  $ \mathcal{F}(\cdot)$  using language ($l$), audio ($a$), and visual ($v$) modalities. This conventional training procedure is represented as $\hat{y}_{i} = \mathcal{F}(l, a, v)$, where $\hat{y}_{i} \in \mathbb{R}$ is a  sentimental intensity variable. Aligned with previous mainstream works \cite{tsai2019multimodal,Hazarika2020misa,sun2022cubemlp,yu2021learning,han2021improving}, we regard MSA as a regression task to ensure a fair comparison. 

\noindent \textbf{Cause-Effect Look at MSA.}
To diagnose the causal relationships among variables, we formulate a causal graph
to summarize the MSA framework. Here, we represent a random variable as a capital letter (\eg, $L$), and denote its observed value as a lowercase letter (\eg, $l$). Theoretically, a causal graph $\mathcal{G} = \{ \mathcal{N}, \mathcal{E} \}$ is considered a directed acyclic graph, which represents how a set of variables $\mathcal{N}$ convey causal effects through the causal links $\mathcal{E}$. It provides an intuitive reference to causal correlations for counterfactual analysis \cite{tang2020unbiased, niu2021counterfactual} and causal intervention \cite{pearl2009causal,glymour2016causal}. In \cref{graph}(a), there are six variables in MSA causal graph, including language modality $L$, audio modality $A$, visual modality $V$, multimodal representation $M$, harmful confounders $Z$, and prediction $Y$. From causal theories \cite{pearl2000models, pearl2009causal}, the adverse dataset biases as the confounders to ``poison''  models. All causal relationships among them are explained as follows:

\noindent $\blacktriangleright$ \textbf{Link} $ \bm{(L, A, V)} \rightarrow \bm{M}  \rightarrow \bm{Y}$. Following biased learning \cite{Hazarika2020misa, tsai2019multimodal}, the causal path $(L, A, V) \rightarrow M $ indicates that the multimodal inputs $(L, A, V)$ produce the final multimodal representation $M$ through MSA models $ \mathcal{F}(\cdot)$ :
\begin{equation}
m = \xi_{M}(L = l, A = a, V = v),
\end{equation}
where $\xi_{M}(\cdot)$ is a fusion strategy that depends on different models (\eg, Transformer \cite{Hazarika2020misa} or concatenation \cite{tsai2019multimodal}). Subsequently, the link $ M \rightarrow Y$ reflects that MSA models estimate the desired prediction $Y$ based on pure $M$.

\noindent $\blacktriangleright$ \textbf{Link} $ \bm{M} \leftarrow \bm{Z} \rightarrow \bm{Y}$.
According to~\cite{pearl2000models}, 
the confounders $Z$ are the common cause of $M$ and $Y$. The dataset biases follow the backdoor causal path $ M \leftarrow Z \rightarrow Y$ to establish spurious associations to prevent the models from pursuing true causal effects, which we should eliminate.

Without loss of generality, 
the nodes $(L, A, V)$ are omitted for simplicity since they are not directly affected by $Z$. The new causal graph is illustrated in \cref{graph}(b). Existing models rely on the likelihood $P(Y|M)$ following the new graph. This process is formulated via the Bayes rule \cite{wang2020visual}:
\begin{equation}
\mathcal{F}(m) =  P(Y|M) = \sum_{z}^{}P(Y|M, z)P(z|M),
\end{equation}
where $z$ is any confounder caused by the label or context bias. In this case, MSA models would invariably focus on the statistical shortcut or spurious correlation to perform biased predictions, significantly limiting their performance.
To remove the detrimental effect caused by $z$, our insight is to embrace backdoor adjustment \cite{pearl2009causal}, \ie, predicting an actively intervened outcome via the $do$-operator \cite{glymour2016causal}. As a typical causal intervention, $do(\cdot)$ prevents the effect of parent nodes that cause variables from the non-causal direction, \ie, $Z\rightarrow M$.
As shown in \cref{graph}(c), the intervention cuts the causal path from $Z$ to $m$, \ie, $m$ is no longer affected by $Z$. In practice, we intervene $m$ based on counterfactual embeddings under different scenarios to purify the pure label and context biases in ~\cref{3.3,3.4}.

\subsection{Label Bias Purification}
\label{3.3}
 As \cref{model}(a) shows, the unbalanced label distribution (\ie, ``positive'' dominates the training data over ``negative'') misleads MSA models to establish non-causal associations between the input samples and the positive category. In this case, MSA models would give predictions based on statistical shortcuts even though the contents of the multimodal testing samples are not observed~\cite{grice1961symposium}. To implement the theoretical $do(\cdot)$ intervention, we utilize $\hat{m}$ to denote the imagined counterfactual multimodal representation. The intervention-based counterfactual outcome is as follows:

\begin{equation}
 \begin{split}
    P(Y|do(M)) = P(Y|M = \hat{m} ) = \mathcal{F}(\hat{m}),\\
     \hat{m} = \xi_{M}(L = \hat{l}, A = \hat{a}, V = \hat{v}). \\
 \end{split}
 \end{equation}
Here $\hat{l}, \hat{a}$, and $\hat{v}$ represent the no-treatment condition where $l, a$, and $v$ are not given.
As MSA models cannot ``see'' any multimodal inputs after the intervention, the counterfactual output $\mathcal{F}(\hat{m})$ actually reflects the purely adverse effect from the trained models, \ie, the label bias captured by $Z$. Considering that neural network models cannot deal with void inputs, we utilize average features over the entire training set as counterfactual embeddings for different modalities:
\begin{equation}
\hat{l} = \frac{1}{N} \sum_{i}^{N} l_{i}, \hat{a} = \frac{1}{N} \sum_{i}^{N} a_{i}, \hat{v} = \frac{1}{N} \sum_{i}^{N} v_{i},
 \end{equation}
where $N$ is the number of training samples. Empirically, the average embedding usually produces a distribution similar to the ideal bias and forces the models to decouple the outcome of the harmful bias as humans do \cite{tang2020unbiased}. 

\subsection{Context Bias Purification}
\label{3.4}

Motivated by human decision-making that combines exogenous and endogenous reasons \cite{van2015cognitive}, language utterances can be summarized in the \emph{main content words} and \emph{context words}. The main content words provide valuable semantics clues (\eg, emotionally-beneficial semantics). Conversely, the context words (\eg, stop words or a part of adjectives) as the confounders trick the models into focusing on spurious correlations between semantically-unimportant contexts and specific categories (\eg, \emph{good} $\leftrightarrow$ \emph{positive} mapping). To this end, we use $\tilde{m}$ to achieve another counterfactual outcome with only context words:
\begin{equation}
 \begin{split}
    P(Y|do(M)) = P(Y|M = \tilde{m} ) = \mathcal{F}(\tilde{m}),\\
     \tilde{m} = \xi_{M}(L = \tilde{l}, A = \breve{a}, V = \breve{v}). \\
 \end{split}
 \end{equation}
Here $\tilde{l}$ denotes counterfactual word embedding where the main content words are masked. The mask operation process is as follows:
\begin{align}
\label{eq:mask_operation}
\forall w_{j} \in \tilde{l}, \left\{\begin{matrix}
w_{j} \longleftarrow \mbox{[MASK]} & \text{if}  \enspace w_{j} \in l_{\text{content}}, \\
 w_{j} \longleftarrow w_{j}& \text{if}   \enspace w_{j} \in l_{\text{context}},
\end{matrix}\right.
\end{align}
where [MASK] symbol is a special token to mask a single word $w_j$.
Meanwhile, $\breve{a}$ and $\breve{v}$ denote unseen empty embeddings, \emph{a.k.a.}, zero feature embeddings. In this situation, MSA models could only rely on visible context words to make bias-based predictions. Essentially, the counterfactual outcome $\mathcal{F}(\tilde{m})$ reflects the pure side effect from the trained vanilla models and word-level harmful context bias.

\subsection{Bias Elimination Strategy}

Thanks to humans' innate counterfactual intuition~\cite{niu2021counterfactual,qian2021counterfactual}, we can wisely reveal actual causal effects among variables in biased observations rather than superficial connections.
The human inference process for unbiased decisions is essentially achieved by comparing factual and counterfactual outcomes \cite{pearl2009causality}.
To block the transfer of biases from the training data to the inference process, we imitate such human intuition to introduce an operationally simple yet empirically powerful subtraction operation (\ie, bias elimination strategy).
The debiased prediction via the strategy is as follows:
\begin{equation}
    \aleph (m) = \mathcal{F}(m) - ( \hat{\lambda} \,  \mathcal{F}(\hat{m}) + \tilde{\lambda} \, \mathcal{F}(\tilde{m})),
\end{equation}
where $\mathcal{F}(m)$ and $ \aleph(m)$  correspond to the traditional factual prediction and counterfactual prediction, respectively. $ \mathcal{F}(\hat{m}) $  and $\mathcal{F}(\tilde{m})$  are the label bias and context bias purified from the poisoned models. Two adaptive trade-off parameters, $\hat{\lambda}$ and $\tilde{\lambda}$, are applied to measure the extent of label bias and context bias.
Since different datasets suffer from varying extent of biases, the grid search strategy is utilized on the validation set to estimate the extent to which the two biases poison the models. We implement the search for $\hat{\lambda}$ and $\tilde{\lambda}$ in a two-dimensional space of a specific interval:
\begin{equation}
    \hat{\lambda}^*, \, \tilde{\lambda}^*=\mathop{\arg\max} \limits_{\hat{\lambda}, \, \tilde{\lambda} \in[\alpha, \beta] } \Phi_{\mathcal{D}}\left(\aleph(m|\hat{\lambda}, \tilde{\lambda})\right),
\end{equation}
where $[\alpha$, $\beta]$ is the search interval. $\Phi (\cdot)$ is a function used for calculating a specific metric that measures the model's performance on the validation set $\mathcal{D}$. 
The evaluation metric is the weighted F1-score, which is the balanced harmonic mean of precision and recall and can excellently reflect the extent of the dataset biases, especially for the imbalanced data.
To reduce the invalid computational overhead during the bias elimination process, we employ a coarse-to-fine grid search strategy to perform a search by gradually narrowing the search interval and step size. As two dataset-level parameters, they are searched only once for each validation set and can be used in inference for all testing data.

\section{Experiments}
\subsection{Datasets and Evaluation Metrics}

\textbf{Datasets.}
Here, we conduct experiments on two different scales of datasets that show significant label and context biases \cite{sun2022counterfactual}.
\textbf{MOSI}~\cite{zadeh2016multimodal} is a realistic dataset comprising 2,199 opinion video clips collected from YouTube. There are 1,284, 229, and 686 video clips in train, valid, and test data, respectively. \textbf{MOSEI}~\cite{zadeh2018multimodal} benchmark contains 23,453 annotated video segments from over 1,000 speakers and 250 topics. There are a total of 16,326, 1,871, and 4,659 video segments in training, validation, and testing sets, respectively. Each sample has a label for both datasets from -3 (strongly negative) to +3 (strongly positive).

\noindent \textbf{Evaluation Metrics.} 
 Following previous works~\cite{lv2021progressive,liang2021attention}, we leverage various metrics to evaluate the MCIS framework's performance, including seven-class classification accuracy (\textbf{Acc-7}) meaning the proportions of correct predicted scores in seven intervals from -3 to +3, binary classification accuracy (\textbf{Acc-2}), and the weighted \textbf{F1} score computed for positive/negative classification results.

\subsection{Model Zoo}
To fully evaluate the effectiveness of MCIS across different methods, we select five representative and reproducible state-of-the-art (SOTA) models. Concretely, \textbf{MulT}~\cite{tsai2019multimodal}
learns element correlations among modalities via paired cross-modal attention interactions.
\textbf{MISA}~\cite{Hazarika2020misa}
projects each modality into two distinct subspaces to learn the discrepancy and consistency across modalities separately.
\textbf{CubeMLP}~\cite{sun2022cubemlp} utilizes three independent multi-layer perceptron units for feature-mixing on three axes.
\textbf{MMIM}~\cite{han2021improving} maximizes the mutual information during multimodal fusion to maintain task-related information.
\textbf{DMD}~\cite{li2023decoupled} introduces cross-modal distillations to facilitate the transfer of informative semantics from strong to weak modalities.

\subsection{Implementation Details}
\noindent\textbf{Feature Extraction.} Following the original protocols of the models above, the audio and visual features are provided by MOSI and MOSEI. The language embeddings are extracted by the pre-trained BERT~\cite{devlin2018bert}, whether fine-tuning depends on the vanilla settings of different methods.
Moreover, we employ the Python NLTK toolkit to tokenize sentences into word lists and then extract the main content words that may affect the semantics in the transcripts.
The average mask ratio of the main content words is 68.96\%. 
For the grid search strategy, the search step and search interval are 0.5 and [-2.0, 2.0] in the coarse search process. In the fine search process, the search step is 0.1, while the search interval depends on the results of the coarse search process.

\noindent\textbf{Experimental Setup.} We re-implement these five SOTA models based on the public codebase and combine them with our MCIS framework. All models are reproduced on NVIDIA Tesla V100 GPUs. For impartiality, the training settings of these models (\eg, loss function, batch size, learning rate strategy, and other hyper-parameters) are consistent with the details reported in original papers.

\begin{table}[t]
    \centering
    \begin{minipage}[t]{0.45\textwidth}
        \centering
\caption{Comparison results on the MOSI testing set. All models use the BERT-based word embedding. $\dagger$: reproduced results from public code with hyper-parameters provided in original papers. The improved results are marked in \textbf{bold}. }
\resizebox{\linewidth}{!}{%
\begin{tabular}{c|ccc}
\toprule
Models        & Acc-7 (\%)                           & Acc-2 (\%)                           & F1 (\%)                              \\ \midrule
TFN~\cite{zadeh2017tensor}            & 34.9                                 & 80.8                                 & 80.7                                 \\
LMF~\cite{liu2018efficient}           & 33.2                                 & 82.5                                 & 82.4                                 \\
MFM~\cite{tsai2018learning}            & 35.4                                 & 81.7                                 & 81.6                                 \\
ICCN~\cite{sun2020learning}           & 39.0                                   & 83.0                                   & 83.0                                   \\
MAG-BERT~\cite{rahman2020integrating}       & 43.6                                 & 84.4                                 & 84.6                                 \\
FDMER~\cite{yang2022disentangled}          & 44.1                                 & 84.6                                 & 84.7                                 \\
Self-MM~\cite{yu2021learning}        & 45.8                                 & 84.8                                 & 84.9                                 \\ \midrule
MulT$^{\dagger}$(ACL'19)~\cite{tsai2019multimodal}           & 42.6                                 & 84.1                                 & 83.9                                 \\  \rowcolor{red!3}
MulT + \textbf{MCIS}    & \textbf{43.5}                        & \textbf{85.5}                        & \textbf{85.2}                        \\
MISA$^{\dagger}$(ACM MM'20)~\cite{Hazarika2020misa}           & 42.1                                 & 82.3         & 82.6          \\ \rowcolor{red!3}
MISA + \textbf{MCIS}    & 42.0                                 & \textbf{83.7} & \textbf{84.1} \\
MMIM$^{\dagger}$(EMNLP'21)~\cite{han2021improving}           & 46.4                                 & 85.5                                 & 85.4                                 \\ \rowcolor{red!3}
MMIM + \textbf{MCIS}    & \textbf{47.9}                        & \textbf{86.6}                        & \textbf{86.5}                        \\
CubeMLP$^{\dagger}$(ACM MM'22)~\cite{sun2022cubemlp}        & 44.5                                 & 84.7                                 & 84.6                                 \\ \rowcolor{red!3}
CubeMLP + \textbf{MCIS} & \textbf{45.7}                        & \textbf{85.9}                        & \textbf{85.8}                        \\
DMD$^{\dagger}$(CVPR'23)~\cite{li2023decoupled}            & 45.3          & 85.1          & 85.1          \\ \rowcolor{red!3}
DMD + \textbf{MCIS}     & \textbf{46.5} & \textbf{86.3} & \textbf{86.3} \\ \bottomrule
\end{tabular}
}
\label{tab_mosi}
    \end{minipage}
    \quad 
    \begin{minipage}[t]{0.45\textwidth}
        \centering
\caption{Comparison results on the MOSEI testing set. All models use the BERT-based word embedding. $\dagger$: reproduced results from public code with hyper-parameters provided in original papers. The improved results are marked in \textbf{bold}. }
\resizebox{\linewidth}{!}{%
\begin{tabular}{c|ccc}
\toprule
Models        & Acc-7 (\%)                           & Acc-2 (\%)                           & F1 (\%)                              \\ \midrule
TFN~\cite{zadeh2017tensor}           & 50.2                                 & 82.5                                 & 82.1                                 \\
LMF~\cite{liu2018efficient}            & 48.0                                   & 82.0                                   & 82.1                                 \\
MFM~\cite{tsai2018learning}            & 51.3                                 & 84.4                                 & 84.3                                 \\
ICCN~\cite{sun2020learning}           & 51.6                                 & 84.2                                 & 84.2                                 \\
MAG-BERT~\cite{rahman2020integrating}       & 52.7                                 & 84.8                                 & 84.7                                 \\
FDMER~\cite{yang2022disentangled}          & 54.1                                 & 86.1                                 & 85.8                                 \\
Self-MM~\cite{yu2021learning}        & 53.5                                 & 85.0                                   & 84.9                                   \\ \midrule
MulT$^{\dagger}$(ACL'19)~\cite{tsai2019multimodal}           & 52.3                                 & 82.7                                 & 82.5                                 \\ \rowcolor{red!3}
MulT + \textbf{MCIS}    & \textbf{54.1}                        & \textbf{84.3}                        & \textbf{84.0}                          \\
MISA$^{\dagger}$(ACM MM'20)~\cite{Hazarika2020misa}           & 52.1                                 & 84.4          & 84.2          \\ \rowcolor{red!3}
MISA + \textbf{MCIS}    & \textbf{53.6}                        &  \textbf{85.8} & \textbf{85.7} \\
MMIM$^{\dagger}$(EMNLP'21)~\cite{han2021improving}           & 53.1                                 & 85.1                                 & 85.0                                 \\ \rowcolor{red!3}
MMIM + \textbf{MCIS}    & \textbf{54.5}                        & \textbf{86.7}                        & \textbf{86.6}                        \\
CubeMLP$^{\dagger}$(ACM MM'22)~\cite{sun2022cubemlp}        & 52.7                                 & 84.2                                 & 83.7                                 \\ \rowcolor{red!3}
CubeMLP + \textbf{MCIS} & \textbf{54.2}                        & \textbf{86.2}                        & \textbf{85.9}                        \\
DMD$^{\dagger}$(CVPR'23)~\cite{li2023decoupled}            &  53.9          & 85.6          & 85.5          \\ \rowcolor{red!3}
DMD + \textbf{MCIS}     & \textbf{55.2} & \textbf{87.3} &  \textbf{87.1} \\ \bottomrule
\end{tabular}
}
\label{tab_mosei}
    \end{minipage}
\end{table}

\subsection{Comparison with State-of-the-art Methods}
We compare the MCIS-based models with recent competitive methods, including TFN~\cite{zadeh2017tensor}, LMF~\cite{liu2018efficient}, MFM~\cite{tsai2018learning}, ICCN~\cite{sun2020learning}, MAG-BERT~\cite{rahman2020integrating}, FDMER~\cite{yang2022disentangled}, and Self-MM~\cite{yu2021learning}. The results on MOSI and MOSEI are reported in Tables \ref{tab_mosi}\&\ref{tab_mosei}. The key observations are as follows. (\textbf{\rmnum{1}}) The models with MCIS significantly and consistently outperform the vanilla versions by large margins on most evaluation metrics for both datasets. 
In particular, the MCIS-based MMIM~\cite{han2021improving} achieves new SOTAs with the Acc-7/Acc-2/F1 scores of 47.9\%/86.6\%/86.5\% on MOSI.
Thanks to MCIS, the distillation-based DMD~\cite{li2023decoupled} yields the best results on MOSEI with affluent improvements of 1.3\%, 1.7\%, and 1.6\% on these three metrics.
The performance gains across methods with different representation learning patterns~\cite{Hazarika2020misa,han2021improving,li2023decoupled} and fusion strategies~\cite{tsai2019multimodal,sun2022cubemlp} confirm the usefulness and generalizability of our framework.

(\textbf{\rmnum{2}}) Compared to existing models that obtain inadequate results (average about 0.54\%$\sim $1.26\% gain across all metrics) via complex structures and numerous parameters \cite{yang2022disentangled,rahman2020integrating,sun2020learning,tsai2019multimodal,yu2021learning,tsai2018learning,li2023decoupled,zadeh2017tensor,liu2018efficient}, MCIS can easily achieve superior improvements (average about 0.94\%$\sim$ 1.76\% gain across all metrics) by removing harmful biases only at the inference phase in a \textit{parameter-free manner}. In practice, our framework is cost-effective compared to training a new SOTA model from scratch since the time overhead is reduced by about \textit{26 times} on average. The better results show that these biases are the ignored ``culprits'' and the importance of counterfactual debiasing.

(\textbf{\rmnum{3}}) Furthermore, we find that the MCIS-based models provide better improvements on MOSEI (average about 1.63\% gain across models) than on MOSI (average about 1.16\% gain across models). The phenomenon potentially derives from extensive data samples in the large-scale dataset beneficial to trained models preserving the two biases that obey the ideal distribution, thus facilitating MCIS to purify and mitigate the adverse effects more effectively.

\begin{table*}[t]
\renewcommand{\arraystretch}{1.2}
\centering   
\caption{ Ablation study results of different dataset biases. We provide comprehensive results for five MCIS-based SOTA models on the MOSEI testing set. Similar trends are also observed on the MOSI. ``w/o'' is short for the without. }
\resizebox{\linewidth}{!}{%
\begin{tabular}{c|ccc|ccc|ccc|ccc|ccc}
\toprule
\multirow{2}{*}{Designs/Mechanisms} & \multicolumn{3}{c|}{MulT + \textbf{MCIS}}                                                                         & \multicolumn{3}{c|}{MISA + \textbf{MCIS}}                                                                           & \multicolumn{3}{c|}{MMIM + \textbf{MCIS}}                                                                           & \multicolumn{3}{c|}{CubeMLP + \textbf{MCIS}}                                                                        & \multicolumn{3}{c}{DMD + \textbf{MCIS}}                                                                            \\ \cline{2-16} 
                         & \multicolumn{1}{c}{Acc-7}         & \multicolumn{1}{c}{Acc-2}         & \multicolumn{1}{c|}{F1}          & \multicolumn{1}{c}{Acc-7}         & \multicolumn{1}{c}{Acc-2}         & \multicolumn{1}{c|}{F1}            & \multicolumn{1}{c}{Acc-7}         & \multicolumn{1}{c}{Acc-2}         & \multicolumn{1}{c|}{F1}            & \multicolumn{1}{c}{Acc-7}         & \multicolumn{1}{c}{Acc-2}         & \multicolumn{1}{c|}{F1}            & \multicolumn{1}{c}{Acc-7}         & \multicolumn{1}{c}{Acc-2}         & \multicolumn{1}{c}{F1}            \\ \midrule
Full Framework           & \multicolumn{1}{c}{\textbf{54.1}} & \multicolumn{1}{c}{\textbf{84.3}} & \multicolumn{1}{c|}{\textbf{84.0}} & \multicolumn{1}{c}{\textbf{53.6}} & \multicolumn{1}{c}{\textbf{85.8}} & \multicolumn{1}{c|}{\textbf{85.7}} & \multicolumn{1}{c}{\textbf{54.5}} & \multicolumn{1}{c}{\textbf{86.7}} & \multicolumn{1}{c|}{\textbf{86.6}} & \multicolumn{1}{c}{\textbf{54.2}} & \multicolumn{1}{c}{\textbf{86.2}} & \multicolumn{1}{c|}{\textbf{85.9}} & \multicolumn{1}{c}{\textbf{55.2}} & \multicolumn{1}{c}{\textbf{87.3}} & \multicolumn{1}{c}{\textbf{87.1}} \\
w/o Label Bias Elimination           & 53.8                              & 83.7                              & 83.5                             & 53.2                              & 85.3                              & 85.2                               & 54.2                              & 86.0                                & 85.9                               & 54.0                                & 85.7                              & 85.5                               & 54.8                              & 86.7                              & 86.6                              \\
w/o Context Bias Elimination         & 52.8                              & 83.2                              & 83.1                             & 52.5                              & 84.7                              & 84.7                               & 53.3                              & 85.5                              & 85.4                               & 53.2                              & 84.8                              & 84.5                               & 54.2                              & 86.1                              & 85.9                              \\
w/o  Grid Search Strategy       & 52.6                              & 83.0                              & 82.8                             & 51.5                              & 83.8                              & 83.6                               & 52.3                              & 84.4                              & 84.2                               & 52.8                              & 84.4                                & 84.0                               & 53.7                              & 86.0                                & 85.7                              \\ \bottomrule
\end{tabular}
}
\label{tab_abl1}
\end{table*}

\begin{table}[t]
\setlength{\tabcolsep}{2pt}
\centering   
\caption{Ablation study results of multimodal counterfactual embeddings in the label bias. ``L/A/V/RCE'' stands for language, audio, visual, and random counterfactual embeddings, respectively.}
\resizebox{0.7\linewidth}{!}{%
\begin{tabular}{c|c|ccccc}
\toprule
Models                         & Metrics & Full     & w/o LCE & w/o ACE       & w/o VCE       & w/ RCE \\ \midrule
\multirow{2}{*}{MulT~\cite{tsai2019multimodal} + \textbf{MCIS}}    & Acc-2 (\%)   & \textbf{84.3} & 83.9    & 84.2          & 84.1          & 83.6    \\
                                & F1 (\%)      & \textbf{84.0} & 83.7    & 83.9 & 83.8          & 83.3    \\ \midrule
\multirow{2}{*}{MISA~\cite{Hazarika2020misa} + \textbf{MCIS}}    & Acc-2 (\%)   & \textbf{85.8} & 85.4    & 85.6          & 85.7          & 85.1    \\
                                & F1 (\%)      & \textbf{85.7} & 85.4    & 85.5          & 85.6          & 85.0      \\ \midrule
\multirow{2}{*}{MMIM~\cite{han2021improving} + \textbf{MCIS}}    & Acc-2 (\%)   & \textbf{86.7} & 86.2    & 86.5          & 86.4          & 85.8    \\
                                & F1 (\%)      & \textbf{86.6} & 86.1    & 86.5          & 86.2          & 85.7    \\ \midrule
\multirow{2}{*}{CubeMLP~\cite{sun2022cubemlp} + \textbf{MCIS}} & Acc-2 (\%)   & \textbf{86.2} & 85.8    & 86.0            & 86.1          & 85.5    \\
                                & F1 (\%)      & \textbf{85.9} & 85.6    & 85.7          & \textbf{85.9} & 85.1    \\ \midrule
\multirow{2}{*}{DMD~\cite{li2023decoupled} + \textbf{MCIS}}     & Acc-2 (\%)   & \textbf{87.3} & 86.8    & 87.0            & 87.1          & 86.5    \\
                                & F1 (\%)      & \textbf{87.1} & 86.7    & 86.8          & 87.0            & 86.3    \\ \bottomrule
\end{tabular}
}
\label{tab_abl2}
\end{table}

\subsection{Ablation Studies}
We perform systematic ablation studies using the MCIS-based models on MOSEI. Comprehensive experiments aim to evaluate the different designs and mechanisms in the proposed MCIS.

\noindent\textbf{Analysis of Different Dataset Biases.} 
\Cref{tab_abl1} provides investigations of two types of bias eliminations and grid search strategy (GSS).
(\textbf{\rmnum{1}}) Firstly, the label and context bias eliminations are retained separately to verify the effect of the distinct biases. The gain drops for all metrics reveal that it is indispensable to simultaneously remove statistical shortcuts and spurious correlations. The core explanation is that the purified label bias provides a sample-agnostic global offset and the purified context bias provides utterance-specific local offsets to correct for the predicted space, allowing the trained models to sidestep the interference of harmful biases in the observed data.
(\textbf{\rmnum{2}}) Another finding is that the impact of context bias is more severe than label bias, implying that misleading or unfair context words more easily mislead the trained models. This observation provides pertinent evidence for the dominance of language modality in MSA \cite{tsai2019multimodal, wu2021text}. (\textbf{\rmnum{3}}) When our GSS is eliminated (\ie, $\hat{\lambda} = \tilde{\lambda} =1$), all gain degradation indicates that proper mitigation of varying degrees of biases is essential.

\noindent\textbf{Impact of MCE in Label Bias.} 
Multimodal Counterfactual Embeddings (MCE) play an important role in obtaining the intervened outcomes based on the purified biases.
(\textbf{\rmnum{1}}) In practice, we investigate the necessity of Language, Audio, and Visual Counterfactual Embeddings (L/A/VCE) separately. From the decreased results in \Cref{tab_abl2}, the incomplete counterfactual embeddings (\ie, the absence of whichever of L/A/VCE) would impede the biased models from producing the multimodal representation that benefits from precise intervention, and then fail to imagine the bias-based outcome purely. According to \cref{graph}, the reason could be that $M$ is confounded by the harmful effects of statistical shortcuts conveyed jointly by links to different modalities \ie, $(L, A, V) \rightarrow M $. Therefore, it takes sufficient intervention with each modality to purify the effective label bias.
(\textbf{\rmnum{2}}) Across all MCIS-based models, the worse deterioration is observed with the elimination of LCE. Meanwhile, the impact of A/VCE on gain depends on different models, \eg, removing ACE is less damaging to the performance of MulT and MMIM as well as VCE is slightly impairing MISA, CubeMLP, and DMD.
(\textbf{\rmnum{3}}) Additionally, we empirically provide a candidate assumption that the average features from three modalities are replaced with the Random Counterfactual Embeddings (RCE), which are initialized by random distribution. The poor results are inevitable because random guesses potentially fail to produce a stable distribution similarly distributed with the ideal bias.

\begin{table}[t]
\setlength{\tabcolsep}{5pt}
\renewcommand{\arraystretch}{1.1}
\centering   
\caption{ Ablation study results of multimodal counterfactual embeddings in the context bias. ``Mask'' means the mask operation in \cref{eq:mask_operation}. ``w/'' and ``w/o'' are short for the with and without, respectively. We only report F1 scores for visual clarity. }
\resizebox{0.75\linewidth}{!}{%
\begin{tabular}{c|ccccc}
\toprule
\multirow{2}{*}{Designs} & \multirow{2}{*}{\begin{tabular}[c]{@{}c@{}}MulT~\cite{tsai2019multimodal}\\ + \textbf{MCIS}\end{tabular}} & \multirow{2}{*}{\begin{tabular}[c]{@{}c@{}}MISA~\cite{Hazarika2020misa}\\ + \textbf{MCIS}\end{tabular}} & \multirow{2}{*}{\begin{tabular}[c]{@{}c@{}}MMIM~\cite{han2021improving}\\ + \textbf{MCIS}\end{tabular}} & \multirow{2}{*}{\begin{tabular}[c]{@{}c@{}}CubeMLP~\cite{sun2022cubemlp}\\ + \textbf{MCIS}\end{tabular}} & \multirow{2}{*}{\begin{tabular}[c]{@{}c@{}}DMD~\cite{li2023decoupled}\\ + \textbf{MCIS}\end{tabular}} \\
                         &                                                                        &                                                                        &                                                                        &                                                                           &                                                                       \\ \midrule
Full Framework           & \textbf{84.0}                                                          & \textbf{85.7}                                                          & \textbf{86.6}                                                          & \textbf{85.9}                                                             & \textbf{87.1}                                                         \\ \midrule
w/o Mask                 & 83.2                                                                   & 84.9                                                                   & 85.6                                                                   & 84.7                                                                      & 86.3                                                                  \\
w/ All Mask              & 83.6                                                                   & 85.2                                                                   & 86.1                                                                   & 85.4                                                                      & 86.7                                                                  \\
w/ Random Mask           & 83.3                                                                   & 84.6                                                                   & 85.9                                                                   & 85.0                                                                        & 86.2                                                                  \\
w/o ACE                  & 83.7                                                                   & 85.6                                                                   & 86.5                                                         & 85.8                                                                      & 86.9                                                                  \\
w/o VCE                  & 83.9                                                                   & 85.4                                                                   & 86.4                                                                   & 85.7                                                                      & \textbf{87.1}                                                         \\
w/ RCE                   & 82.8                                                                   & 84.4                                                                   & 85.4                                                                   & 84.3                                                                      & 85.8                                                                  \\ \bottomrule
\end{tabular}
}
\label{tab_abl3}
\end{table}

\begin{figure*}[t]
  \centering
  \includegraphics[width=\textwidth]{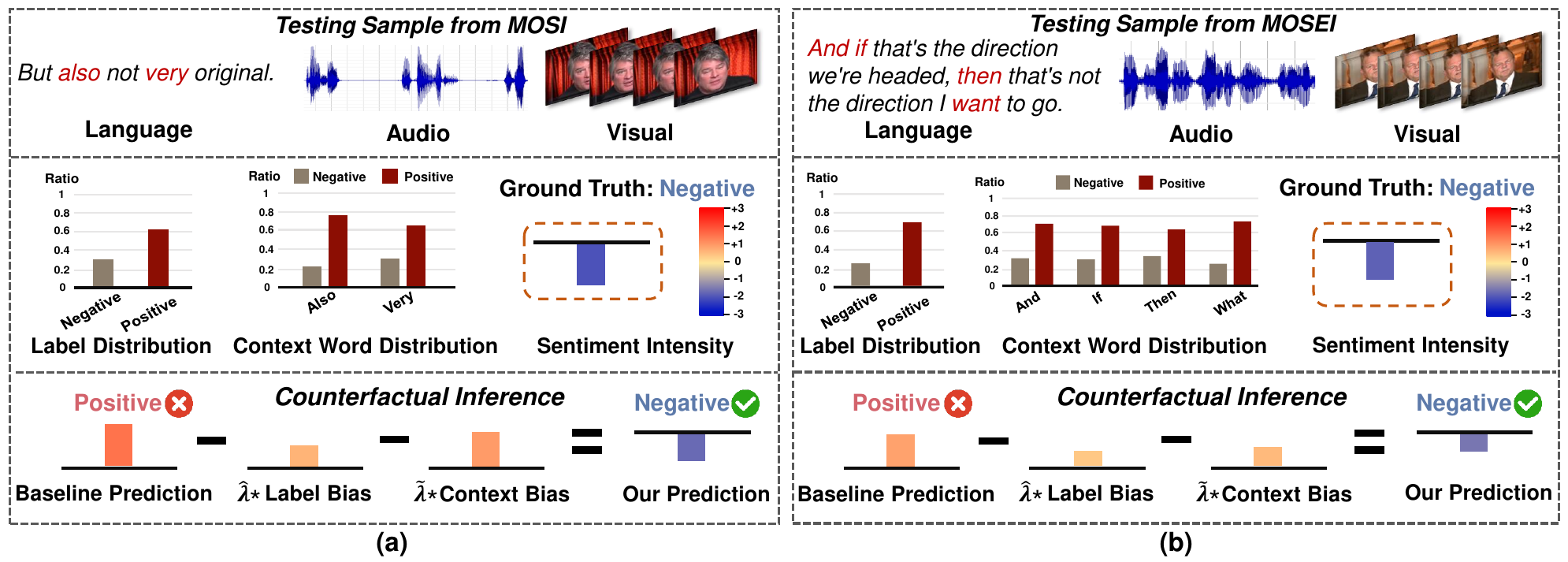}
  \caption{ Case study of counterfactual learning on MOSI and MOSEI.
  We report the binary evaluation results from the DMD~\cite{li2023decoupled} with our MCIS for the intuitive display. Label/Context Word Distribution: the imbalanced distribution of sentiment labels and context words in positive and negative categories comes from the training set.
 }
  \label{case}
\end{figure*}

\noindent\textbf{Impact of MCE in Context Bias.} 
Intuitively, the core of context bias elimination is masking the main content words and forcing the models to focus only on the spurious correlations provided by the context words. To explore this, (\textbf{\rmnum{1}}) we perform the word non-masking (w/o Mask), all masking (w/ All Mask), and random masking (w/ Random Mask) separately before converting the transcripts into word embeddings via the pre-trained BERT in \Cref{tab_abl3}. 
The decreased results in F1 scores confirm three explanations: 
(1) Due to the language modality unavailability in all masking, the Context Bias Elimination (CBE) process does not impact linguistic effects. Despite the bias of vanilla models, the main content words contribute more valuable gains. (2) Instead, CBE purifies the effects of both good semantics and bad bias in non-masking, leading to worse results. (3) The poor results for random masking than for all masking suggest that CBE probably over-eliminates meaningful clues as the main content words dominate.
(\textbf{\rmnum{2}}) Furthermore, the original features of different training samples are retained when ACE and VCE are removed separately.
The most gain drops suggest that our zero feature embedding assumption guarantees a safe estimation for the purification of pure word-level context bias. (\textbf{\rmnum{3}}) As an alternative to L/A/VCE, the worst performance from all metrics with RCE verifies the rationality of the proposed embedding paradigm.

\subsection{Qualitative Analysis}
\noindent\textbf{Case Study of Counterfactual Learning.} 
\cref{case} shows a counterfactual example from each testing set on MOSI and MOSEI, respectively. Specifically, we provide the sentiment intensity scores of positive/negative evaluation results from vanilla DMD, two types of counterfactual outputs, and the counterfactual predictions.
The corresponding label and context word distributions for the display samples intuitively show the presence of the dataset biases.
Evidently, MCIS corrects the baseline predictions and gives reasonable sentiment polarities. Taking Case 1 (\cref{case}(a)) as an example, the vanilla model obtains a falsely positive polarity, which is misled by the dataset biases. According to the two counterfactual outputs corresponding to the purified biases, the biased baseline results suffer from two deleterious effects: (1) the statistical shortcut caused by the large proportion of ``positive'' labels; (2) the spurious correlation between the context words (\eg, ``also'', ``very'') and ``positive'' category. Thanks to the proposed MCIS, we can empower the model to think twice and make unbiased predictions by comparing factual and counterfactual outcomes. 

 \begin{figure}[t]
  \centering
  \includegraphics[width=\linewidth]{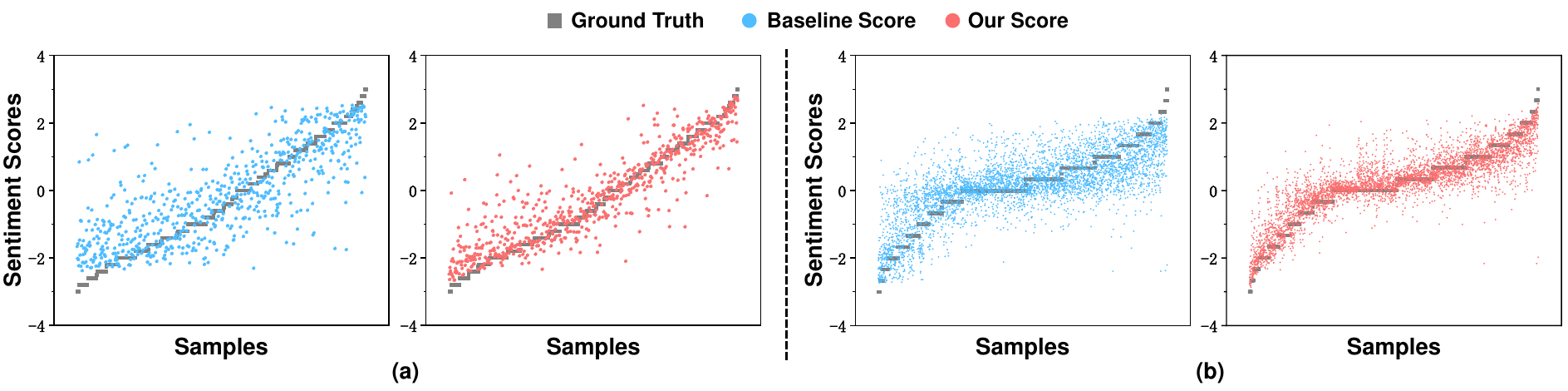}
  \caption{Distribution differences of sentiment scores for the testing set (sorted) on (a) MOSI and (b) MOSEI. 
  The blue dots represent the predicted scores from the baseline DMD~\cite{li2023decoupled}, while the red dots represent the predicted scores from the MCIS-based DMD. The more compact the distribution of predicted sentiment scores and ground truths, the better the model performance.
  }

  \label{case2}
\end{figure}

\noindent\textbf{Distribution Differences of Sentiment Scores.} The distribution differences of sentiment scores on MOSI and MOSEI testing sets are displayed in \cref{case2}(a) and \cref{case2}(b), respectively. 
\textbf{(i)} Macroscopically, the predicted score distribution of the MCIS-based model is more compact with the ground truth distribution, indicating that MCIS can effectively correct prediction errors around ground truths. 
\textbf{(ii)} In practice, our framework mitigates the overall prediction gap caused by samples with outlier-predicted scores while maintaining correct predictions for most samples.
For instance, the MCIS-based DMD successfully corrects about 90\% and 93\% of the predicted sentiment scores in samples with changes in sentiment polarities on MOSI and MOSEI.
\textbf{(iii)} Microscopically, MCIS differs in its debiasing effect on different samples, depending on the misleading extent of the context words in the samples. In short, our method contributes to a meaningful step towards the unbiased estimation of existing models.

\section{Conclusion}
In this paper, we investigate and disentangle the dataset biases that have long poisoned MSA models from a causal inference perspective.
As a model-agnostic causality-based framework, the proposed MCIS eliminates the detrimental effects caused by these biases via imitating human counterfactual intuition. Comprehensive experiments demonstrate that the MCIS-based models achieve better performance than their biased counterparts. 

% \noindent \textbf{Limitation.} Following existing MSA datasets \cite{zadeh2018multimodal,zadeh2016multimodal}, our framework relies on the setting of complete input modalities. When the MCIS-based models are applied to realistic scenarios with missing modalities, both types of counterfactual operations may not adequately capture pure biases, causing sub-optimal gains.

\noindent \textbf{Future Work.}
We plan to equip MCIS with modality reconstruction techniques to cope with potential modality missingness in realistic applications. 

\noindent{\textbf{Acknowledgement.}} 
This work is supported in part by the National Key R\&D Program of China under
Grant 2021ZD0113503 and in part by the Shanghai Municipal Science and Technology Major Project
under Grant 2021SHZDZX0103.

% ---- Bibliography ----
%
% BibTeX users should specify bibliography style 'splncs04'.
% References will then be sorted and formatted in the correct style.
%
\bibliographystyle{splncs04}
\bibliography{main}
\end{document}